# Stiffness matrix of manipulators with passive joints: computational aspects

Klimchik A., Pashkevich A., Caro S., and Chablat D.

*Abstract*—The paper focuses on stiffness matrix computation for manipulators with passive joints, compliant actuators and flexible links. It proposes both explicit analytical expressions and an efficient recursive procedure that are applicable in the general case and allow obtaining the desired matrix either in analytical or numerical form. Advantages of the developed technique and its ability to produce both singular and non-singular stiffness matrices are illustrated by application examples that deal with stiffness modeling of two Stewart-Gough platforms.

## I. INTRODUCTION

IN many applications, manipulator stiffness becomes one of the most important performance measures of a robotic system. To evaluate stiffness properties, several methods can be applied such as Finite Element Analysis, Matrix Structural Analysis and Virtual Joint Modeling (VJM) [1-12], where the last one is the most attractive in robotic domain since it operates with an extension of the traditional rigid model that is completed by a set of compliant virtual joints (localized springs), which describe elastic properties of the links, joints and actuators. This paper contributes to the VJM-based technique and focuses on some particularities of the manipulators with passive joints.

For conventional serial manipulators (without passive joints, whose stiffness is equal to zero), the VJM approach yields rather simple analytical presentation of the desired stiffness matrix $\mathbf{K}_C$. Relevant expression $\mathbf{K}_C = \mathbf{J}_\theta^{-T} \mathbf{K}_\theta \mathbf{J}_\theta^{-1}$ can be found in the work of Salisbury [1] and other authors. Here, the matrix $\mathbf{K}_\theta$ aggregates the stiffness coefficients of all elastic virtual joints, and $\mathbf{J}_\theta$ is the corresponding kinematic Jacobian. Further, this result was extended by Gosselin for the case of parallel manipulators (with numerous passive joints) assuming that the manipulator structure is not over-constrained [2]. For more general case, that includes overconstrained architectures, a solution was proposed in our previous work [13], but the developed technique requires rather intensive numerical computations related to high-dimensional matrix inversion. This work focuses on reduction of the computational complexity by means of analytical inversion of some sub-matrices and application of dedicated recursive procedures.

It is also worth mentioning that some previous works [14] propose (or at least discuss) a trivial solution of the considered problem, which deals with a straightforward modification of the stiffness matrix $\mathbf{K}_\theta$, in accordance with the passive joint type and geometry (corresponding rows and columns are simply set to zero). However, as follows from our study, this approach gives true results if (and only if) the matrix $\mathbf{K}_\theta$ is diagonal. It is clear that it is not valid in the general case where there is a coupling between different types of the elementary virtual springs presented by non-diagonal elements of $\mathbf{K}_\theta$. Non-triviality of this problem is clearly confirmed by a motivation example presented in web-appendix of this paper [15], which deals with a single passive joint.

## II. PASSIVE JOINTS IN A SERIAL CHAIN

In contrast to conventional serial manipulators, whose kinematics does not include passive joints and assures full controllability of the end-effector, parallel manipulators include a number of under-actuated serial chains that are mutually constrained by special connection to the base and to the end-platform. Let us derive an analytical expression for the stiffness matrix of such kinematic chain taking into account influence of the passive joints.

The kinematic chain under study (Fig.1) consists of a fixed base, a series of flexible links, a moving platform, and a number of actuated or passive joints separating these elements. Following the methodology proposed in our previous publications [13], a relevant VJM model may be presented as a sequence of rigid links separated by passive joints and six-dimensional virtual springs describing elasticity of the links and actuators.

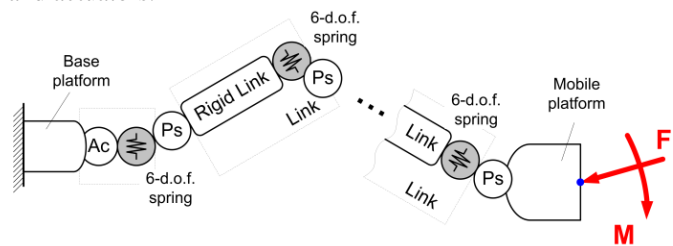

**Fig. 1.** The VJM model of a general serial chain (Ps – passive joint, Ac – actuated joint)

It can be proven that the static equilibrium equations of this

Manuscript received September 07, 2011. The work presented in this paper was partially funded by the Region "Pays de la Loire" (France) and by the ANR, France (project COROUSSO).

A. Klimchik and A. Pashkevich are with Ecole des Mines de Nantes, 4 rue Alfred-Kastler, Nantes 44307, France and with Institut de Recherches en Communications et en Cybernetique de Nantes (IRCCyN), 44321 Nantes, France (phone: Tel.+33-251-85-83-00; fax.+33-251-85-83-49; e-mail: alexandr.klimchik@mines-nantes.fr, anatol.pashkevich@mines-nantes.fr).

S. Caro and D. Chablat are with IRCCyN (e-mail: stephane.caro@irccyn.ec-nantes.fr, damien.chablat@irccyn.ec-nantes.fr).

mechanical system may be written as

$$\mathbf{J}_\theta^T \mathbf{F} = \mathbf{K}_\theta \boldsymbol{\theta}; \quad \mathbf{J}_q^T \mathbf{F} = \mathbf{0}; \quad \mathbf{t} = \mathbf{g}(\mathbf{q}, \boldsymbol{\theta}) \qquad (1)$$

where $\mathbf{J}_q$, $\mathbf{J}_\theta$ are kinematic Jacobians with respect to the passive and virtual joint coordinates $\mathbf{q}$, $\boldsymbol{\theta}$ respectively; $\mathbf{F}$ is the external loading (force and torque), $\mathbf{K}_\theta$ the aggregated stiffness matrix of the virtual springs, the vector $\mathbf{t}$ includes the position and orientation (Cartesian coordinates and Euler angles) of the platform. Using these equations simultaneously and applying the first-order linear approximation under assumption that corresponding values of the external force $\mathbf{F}$ and the coordinate variations are small enough (see [13] for details), one can derive the matrix expression

$$\begin{bmatrix} \mathbf{K}_C & * \\ * & * \end{bmatrix} = \begin{bmatrix} \mathbf{J}_\theta \mathbf{K}_\theta^{-1} \mathbf{J}_\theta^T & \mathbf{J}_q \\ \mathbf{J}_q^T & \mathbf{0} \end{bmatrix}^{-1} \qquad (2)$$

that allows obtaining the desired Cartesian stiffness matrix $\mathbf{K}_C$ numerically, by extracting a $6\times 6$ sub-matrix in upper-left corner of (2).

In spite of apparent simplicity, the above procedure is not convenient for the parametric stiffness analysis that usually relies on analytical expressions. To derive such expression for the matrix $\mathbf{K}_C$, let us apply the blockwise inversion based on the Frobenius formula [16] that allows to present the desired stiffness matrix as

$$\mathbf{K}_C = \mathbf{K}_C^0 - \mathbf{K}_C^0 \mathbf{J}_q (\mathbf{J}_q^T \mathbf{K}_C^0 \mathbf{J}_q)^{-1} \mathbf{J}_q^T \mathbf{K}_C^0. \qquad (3)$$

Here, the first term $\mathbf{K}_C^0 = (\mathbf{J}_\theta \mathbf{K}_\theta^{-1} \mathbf{J}_\theta^T)^{-1}$ is the stiffness matrix of the corresponding serial chain *without passive joints* and the second term defines the stiffness reduction *due to the passive joints*. It worth mentioning that this result is in good agreement with other relevant works [8][17] where $\mathbf{K}_C$ was presented as the difference of two similar components but they were computed in a different way.

Analyzing the latter expression, one can get the following conclusions concerning the stiffness matrix properties:

**Remark 1.** The first term of the expression (3) is non-singular if and only if $rank(\mathbf{J}_\theta) = 6$, i.e. if the VJM model of the chain includes at least 6 independent virtual springs.

**Remark 2.** The second term of the expression (3) is non-singular if and only if $rank(\mathbf{J}_q) = n_q$, where $n_q$ is the number of passive joints.

**Remark 3.** If both terms of (3) are non-singular and $n_q \geq 1$, their difference produces a symmetrical stiffness matrix, which is always singular and $rank(\mathbf{K}_C) = 6 - n_q$.

**Remark 4.** If the matrix $\mathbf{K}_C^0$ of the chain without passive joints is symmetrical and positive-definite, the stiffness matrix of the chain with passive joints $\mathbf{K}_C$ is also symmetrical but positive-semidefinite.

Further simplification can be achieved by applying blockwise inversion to $(\mathbf{J}_q^T \mathbf{K}_C^0 \mathbf{J}_q)^{-1}$, which presents the main computational difficulty in equation (3). Relevant results are summarized in the following proposition.

**Proposition.** If the chain does not include redundant passive joints, expression (3) allows recursive presentation

$$\mathbf{K}_C^{i+1} = \mathbf{K}_C^i - \mathbf{K}_C^i \mathbf{J}_q^i (\mathbf{J}_q^{i\,T} \mathbf{K}_C^i \mathbf{J}_q^i)^{-1} \mathbf{J}_q^{i\,T} \mathbf{K}_C^i; \qquad i = 0,1,2... \qquad (4)$$

where $\mathbf{K}_C^0 = (\mathbf{J}_\theta \mathbf{K}_\theta^{-1} \mathbf{J}_\theta^T)^{-1}$ and sub-Jacobians $\mathbf{J}_q^i$ are extracted from $\mathbf{J}_q$ in arbitrary order (column-by-column, or by groups of columns), superscripts '*i*' and '*i+1*' define iteration number.

**Corollary.** The desired stiffness matrix $\mathbf{K}_C$ can be computed in $n_q$ steps, by sequential application of expression (4) for each column of the Jacobian $\mathbf{J}_q$ (i.e. for each passive joint separately).

These results give convenient computational techniques that will be used below for obtaining the stiffness matrix for parallel manipulators, they are illustrated in Section V. More detailed and formal proof is presented in web-appendix [15].

### III. PASSIVE JOINTS IN A PARALLEL MANIPULATOR

Let us consider now a parallel manipulator, which usually may be presented as a strictly parallel structure of the actuated serial legs connecting the base and the end-platform (Fig. 2) [18]. Using the methodology described in previous section and applying it to each leg, there can be computed a set of Cartesian stiffness matrices $\mathbf{K}_C^{(i)}$ expressed with respect to the same coordinate system but corresponding to different platform points. If initially the chain stiffness matrices were computed in local coordinate systems, their transformation to the global system is performed in standard way [19].

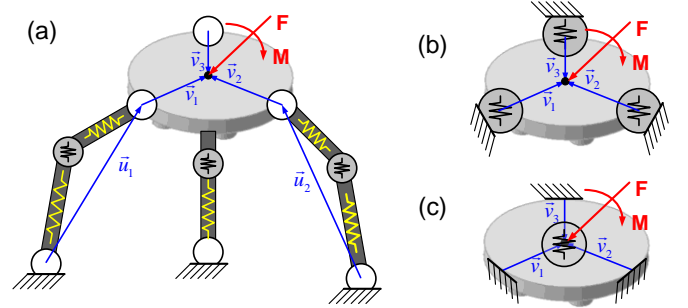

**Fig. 2.** Typical parallel robot (a) and transformation of its VJM models (b, c)

To aggregate these matrices $\mathbf{K}_C^{(i)}$, they must be also re-computed with respect to same reference point of the platform. Assuming that the platform is rigid enough (compared to the legs), this conversion can be performed by extending the legs by a virtual rigid link connecting the end-point of the leg and the reference point of the platform (see Fig. 2 where these extensions are defined by the vectors $\mathbf{v}_i$).

After such extension, an equivalent stiffness matrix of the leg may be expressed using relevant expression for a usual serial chain, i.e. as $\mathbf{J}_v^{(i)-T} \mathbf{K}_C^{(i)} \mathbf{J}_v^{(i)-1}$, where the Jacobian $\mathbf{J}_v^{(i)}$ defines differential relation between the coordinates of the *i*-th leg end-frame and the end-platform reference frame. Hence, using the superposition principle, the final expression for the stiffness matrix of the parallel manipulator can be written as

$$\mathbf{K}_C^{(m)} = \sum_{i=1}^{m} \mathbf{J}_v^{(i)-T} \mathbf{K}_C^{(i)} \mathbf{J}_v^{(i)-1} \tag{5}$$

where $m$ is the number of serial kinematic chains in the manipulator architecture. It is implicitly assumed here that all stiffness matrices (both for the legs and for the whole manipulator) are expressed in the same global coordinate system. Hence, the axes of all virtual springs are parallel to the axes $x$, $y$, $z$ of this system and corresponding Jacobians and their inverses can be easily computed analytically as

$$\mathbf{J}_v^{(i)} = \begin{bmatrix} \mathbf{I}_3 & (\mathbf{v}_i \times) \\ \mathbf{0} & \mathbf{I}_3 \end{bmatrix}_{6\times 6}, \quad \left(\mathbf{J}_v^{(i)}\right)^{-1} = \begin{bmatrix} \mathbf{I}_3 & -(\mathbf{v}_i \times) \\ \mathbf{0} & \mathbf{I}_3 \end{bmatrix}_{6\times 6} \tag{6}$$

where $\mathbf{I}_3$ is the identity matrix of size $3 \times 3$, and $(\mathbf{v}\times)$ is a skew-symmetric matrix corresponding to the vector $\mathbf{v}$.

Therefore, expression (5) allows explicit aggregation of the leg stiffness matrices with respect to any given reference point of the platform. It is worth mentioning that in practice, the matrices $\mathbf{K}_C^{(i)}$ are always singular while there aggregation usually produce a non-singular singular matrix [13].

## IV. COMPUTATIONAL TECHNIQUES

### A. Recursive computations: single-joint decomposition

Let us assume that a current recursion deals with a single passive joint corresponding to the $i$-th column of the Jacobian $\mathbf{J}_q$, which is denoted as $\mathbf{J}_q^i$ and has size $6\times 1$. In this case, the matrix expression $(\mathbf{J}_q^{iT} \mathbf{K}_C^i \mathbf{J}_q^i)^{-1}$ is reduced to the size of $1\times 1$ and the matrix inversion is replaced by a simple scalar division. Besides, the term $\mathbf{K}_C^i \mathbf{J}_q^i$ has size $6\times 1$, so the recursion (4) is simplified to

$$\mathbf{K}_C^{i+1} = \mathbf{K}_C^i - \frac{1}{\mu}\mathbf{u}_i \cdot \mathbf{u}_i^T \quad \text{or} \quad \left[K_{jk}^{(i+1)}\right] = \left[K_{jk}^{(i)}\right] - \frac{1}{\mu}\left[u_j^{(i)} u_k^{(i)}\right] \tag{7}$$

where $\mathbf{u}_i = \mathbf{K}_C^i \mathbf{J}_q^i$ is a $6\times 1$ vector and $\mu = \mathbf{J}_q^{iT} \mathbf{K}_C^i \mathbf{J}_q^i$ is a scalar. It can be also proven that each recursion reduces the rank of the stiffness matrix by 1.

Hence, in the general case, the recursion (4) involves rather non-trivial transformations of $\mathbf{K}_C^i$, different from simple setting to zero a row and/or a column. Let us consider now several specific (but rather typical) cases where the transformation rules are more simple and elegant.

### B. Analytical computations: chains with trivial passive joints

In practice, many parallel robots include kinematic chains for which the passive joint axes are collinear to the axes $x$, $y$ or $z$ of the Cartesian coordinate system. For such architectures, the vector-columns of the Jacobian $\mathbf{J}_q$ include a number of zero elements, so the expressions (7) can be essentially simplified. Let us consider a set of trivial cases where $\mathbf{J}_q^i$ are created from the columns of the identity matrix:

Corresponding passive joints will be further referred to as the 'trivial' ones. It can be easily proven that they cover the following range of the joint geometry:

(i) translational passive joint with arbitrary spatial position (with the joint axis directed along $x$, $y$ or $z$);
(ii) rotational passive joints positioned at the reference point (with the joint axis directed along $x$, $y$ or $z$).

Besides, it is worth to consider additional case-study corresponding to 'quasi-trivial' passive joints:

(iii) rotational passive joints shifted by the distance $L$ with respect to the reference point in the direction either $x$, $y$ or $z$ (with the joint axis directed along $x$, $y$ or $z$).

For the trivial passive joints, assuming that $\mathbf{J}_q^{(p)}$ denotes the vector-column with a single non-zero element in the $p$-th position, the recursive expression (7) for the Cartesian stiffness matrix is simplified to

$$\left[K_{jk}^{(i+1)}\right] = \left[K_{jk}^{(i)}\right] - \left[K_{jp}^{(i)} K_{pk}^{(i)} / K_{pp}^{(i)}\right] \tag{8}$$

that is very simple and can be easily performed analytically. Web-appendix [15] contains a number of examples illustrating stiffness matrix transformations for trivial and quasi-trivial passive joints. They can be used as templates for analytical computations.

## V. APPLICATION EXAMPLES

Let apply now the developed technique to computing of the stiffness matrix for two versions of a general Stewart-Gough platform presented in Fig. 3 [17][20]. It is assumed that in both cases the manipulator base and the moving plate (platform) are connected by six similar extensible legs (Fig. 4) but their spatial arrangements are different:

**Case A**: the legs are regularly connected to the base and platform, with the same angular distance 60°.

**Case B**: the legs are connected to the base and platform in three pairs, with the angular distance of 120° between the mounting.

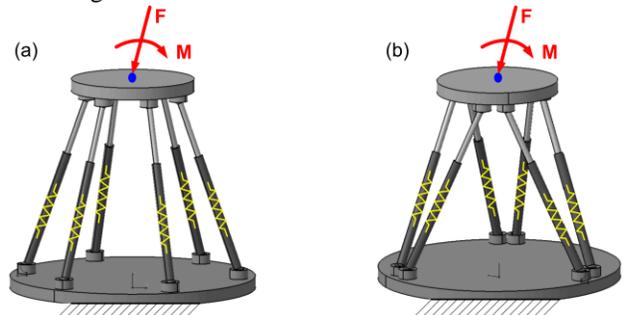

**Fig. 3** Geometry of the Stewart-Gough platforms under study

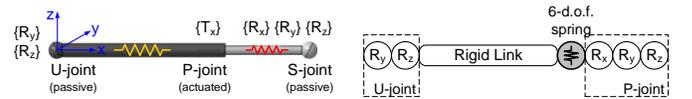

**Fig. 4.** Geometry of the manipulator leg and its VJM model

Using the proposed technique and performing sequentially relevant recursive computations (see web-appendix [15] for details), the desired stiffness matrix of the Gough platform for the both cases can be presented as

$$\mathbf{K}_C = K_{11} \cdot \sum_{i=1}^{6} \begin{bmatrix} \mathbf{u}_i^0 \\ \hline (\mathbf{v}_i \times \mathbf{u}_i^0) \end{bmatrix} \cdot \begin{bmatrix} \mathbf{u}_i^{0\mathrm{T}} & \vdots & (\mathbf{v}_i \times \mathbf{u}_i^0)^{\mathrm{T}} \end{bmatrix} \quad (9)$$

where $K_{11}$ is the corresponding element of $\mathbf{K}_\theta$, the vectors $\mathbf{u}_i$, $\mathbf{v}_i$ describing spatial locations of the legs are computed via the direct kinematics, and $\mathbf{v}_i \times \mathbf{u}_i^0$ is the vector product. Further, after relevant transformations, the desired stiffness matrices for the cases A, and B can be expressed as

$$\mathbf{K}_C^{(A)} = \frac{3K_{11}}{L^2} \begin{bmatrix} d_a^2 & 0 & 0 & \vdots & 0 & rhd_a & 0 \\ 0 & d_a^2 & 0 & \vdots & -rhd_a & 0 & 0 \\ 0 & 0 & 2h^2 & \vdots & 0 & 0 & 0 \\ \hline 0 & -rhd_a & 0 & \vdots & r^2h^2 & 0 & 0 \\ rhd_a & 0 & 0 & \vdots & 0 & r^2h^2 & 0 \\ 0 & 0 & 0 & \vdots & 0 & 0 & 0 \end{bmatrix} \quad (10)$$

and

$$\mathbf{K}_C^{(B)} = \frac{3K_{11}}{L^2} \begin{bmatrix} d_a^2 + Rr & 0 & 0 & \vdots & 0 & rhd_b & 0 \\ 0 & d_a^2 + Rr & 0 & \vdots & -rhd_b & 0 & 0 \\ 0 & 0 & 2h^2 & \vdots & 0 & 0 & 0 \\ \hline 0 & -rhd_b & 0 & \vdots & r^2h^2 & 0 & 0 \\ rhd_b & 0 & 0 & \vdots & 0 & r^2h^2 & 0 \\ 0 & 0 & 0 & \vdots & 0 & 0 & 1.5\,r^2R^2 \end{bmatrix} \quad (11)$$

where $R$, $r$ define location of the leg connection points, $d_a = R - r$; $d_b = R/2 - r$; $L$ is the leg length, $h$ is the vertical distance between the base and the platform. As follows from these expressions, the matrix $\mathbf{K}_C^{(A)}$ is singular and allows "free" rotation of the end-platform around the vertical axis. In contrast, the matrix $\mathbf{K}_C^{(B)}$ is non-singular and the manipulator resists to all external forces/torques applied to the platform. These results are in good agreement with previous research on the Stewart-Gough platforms and confirm efficiency of the developed computational technique for manipulator stiffness modeling [6][20]. Hence, the developed technique allows us obtaining analytical expressions for $\mathbf{K}_C$ rather easily.

## VI. CONCLUSION

For robotic manipulators with passive joints, the stiffness matrices of separate kinematic chains are *singular*. So, the most of existing stiffness analysis methods can not be applied directly. To deal with such architectures in more efficient way, this paper proposes a new approach that allows obtaining both *singular and non-singular* stiffness matrices and which is appropriate for a general case, independent of the type and spatial location of the passive joints. The developed approach is based on the extension of the virtual-joint modeling technique and includes two basic steps which sequentially produce stiffness matrices of separate chains and then aggregate them in a common matrix.

In contrast to previous works, the desired stiffness matrix is presented in an explicit *analytical form*, as a sum of *two terms*. The first of them has traditional structure and describes manipulator elasticity due to the link/joint flexibility, while the second one directly takes into account influence of the passive joints. To simplify analytical computations, it is proposed a *recursive procedure* that sequentially modifies the original matrix in accordance with the geometry of each passive joint.

Advantages of the developed technique are illustrated by application examples that deal with stiffness modeling of two Stewart-Gough platforms. Future work will focus on the extension of these results for the case of parallel manipulators with non-rigid platform and essential external loading.

### ACKNOWLEDGEMENTS

The work presented in this paper was partially funded by the Region "Pays de la Loire", France (project RoboComposite) and by the ANR, France (project COROUSSO).